\documentclass[journal]{IEEEtran}

\ifCLASSINFOpdf
\else
   \usepackage[dvips]{graphicx}
\fi
\usepackage{url}

\usepackage{graphicx}
\usepackage{amsmath}
\usepackage{booktabs}
\usepackage{multirow}
\usepackage{makecell}
\usepackage{threeparttable}
\usepackage[normalem]{ulem} 
\usepackage{array}
\usepackage[nocompress]{cite} 
\usepackage[hidelinks]{hyperref}
\usepackage{amssymb}
\usepackage{booktabs}
\renewcommand{\arraystretch}{1.2} 

\begin{document}

\title{DeRainMamba: A Frequency-Aware State Space Model with Detail Enhancement for Image Deraining}

\author{Zhiliang Zhu, Tao Zeng, Tao Yang, Guoliang Luo and Jiyong Zeng
\thanks{This work was supported in part by the National Natural Science Foundation of China under Grant 62202165 and Grant 62462032,  the Key Research and Development Program of Jiangxi Province under Grant 20223BBE51039 and Grant 20232BBE50020, the Science Fund for Distinguished Young Scholars of Jiangxi Province under Grant 20232ACB212007. }
\thanks{Zhiliang Zhu, Tao Zeng, Tao Yang, and Guoliang Luo are with the School of Information and Software Engineering, East China Jiaotong University, Nanchang 330013, China (e-mail: rj\_zzl@163.com; rj\_zt@ecjtu.edu.cn; rjgc\_yt@163.com; luoguoliang@ecjtu.edu.cn).}
\thanks{Jiyong Zeng is with Lianchuang Electronic Technology Company, Ltd., Nanchang 330013, China (e-mail: zengjiyong7@163.com).}}

\markboth{Journal of \LaTeX\ Class Files, Vol. 14, No. 8, August 2015}
{Shell \MakeLowercase{\textit{et al.}}: Bare Demo of IEEEtran.cls for IEEE Journals}
\maketitle

\begin{abstract}
Image deraining is crucial for improving visual quality and supporting reliable downstream vision tasks. Although Mamba-based models provide efficient sequence modeling, their limited ability to capture fine-grained details and lack of frequency-domain awareness restrict further improvements. To address these issues, we propose DeRainMamba, which integrates a Frequency-Aware State-Space Module (FASSM) and Multi-Directional Perception Convolution (MDPConv). FASSM leverages Fourier transform to distinguish rain streaks from high-frequency image details, balancing rain removal and detail preservation. MDPConv further restores local structures by capturing anisotropic gradient features and efficiently fusing multiple convolution branches. Extensive experiments on four public benchmarks demonstrate that DeRainMamba consistently outperforms state-of-the-art methods in PSNR and SSIM, while requiring fewer parameters and lower computational costs. These results validate the effectiveness of combining frequency-domain modeling and spatial detail enhancement within a state-space framework for single image deraining.

\end{abstract}

\begin{IEEEkeywords}
Image deraining, state space model, frequency-domain modeling, detail enhancement
\end{IEEEkeywords}

\IEEEpeerreviewmaketitle

\section{Introduction}

\IEEEPARstart{I}{mage} deraining aims to restore clear and high-quality images from those captured under rainy conditions, thereby enhancing the visual quality and improving the robustness of downstream tasks such as object detection and segmentation.

Early deep learning-based deraining approaches, mainly relying on convolutional neural networks (CNNs)~\cite{gan2025enhancing,gan2025semamil,li2023ntire,ren2024ninth,wang2025ntire,peng2020cumulative,wang2023decoupling,peng2024lightweight,peng2024towards,wang2023brightness,peng2021ensemble,ren2024ultrapixel,yan2025textual,peng2024efficient,conde2024real,peng2025directing,peng2025pixel,peng2025boosting,he2024latent,peng2024unveiling,he2024dual,he2024multi,pan2025enhance,wu2025dropout,jiang2024dalpsr,ignatov2025rgb,du2024fc3dnet,jin2024mipi,sun2024beyond,qi2025data,feng2025pmq,xia2024s3mamba,pengboosting,suntext,yakovenko2025aim,xu2025camel,wu2025robustgs,li2023cross,li2023survey,li2024efficient1,li2024efficient2,li2025fouriersr,li2025dual,li2025self}, have achieved promising results by learning local features through hierarchical representations. However, due to the limited receptive field of CNNs, these methods often fail to capture global contextual information, resulting in poor detail reconstruction and insufficient degradation removal capability. Consequently, many researchers have focused on long-range modeling using Transformers introduced into image restoration tasks, achieving significant progress~\cite{ref5,ref6,ref7,ref9,zheng2025towards,zheng2024odtrack,zheng2025decoupled,zheng2023toward,zheng2022leveraging,duan2025dit4sr,chen2025mixnet,chen2024towards,wu2025dropout,wu2025adaptive,wu2024rethinking,li2024efficient1,li2025fouriersr,li2024efficient2,li2025dual,li2023cross}. By leveraging the self-attention mechanism, Transformers are capable of modeling long-range dependencies, leading to notable performance improvements. Nevertheless, the inherent quadratic complexity of self-attention with respect to input size severely limits the applicability of Transformers to high-resolution images, particularly in real-time or resource-constrained scenarios~\cite{ref10,ref11,ref43,1,2,3,4,5,6,7,lu2024mace,lu2023tf,lu2024robust,lu2025does,zhou2025dragflow,li2025set,gao2024eraseanything,ren2025all,lu2022copy,yang2025temporal,yu2025visual,gao2025revoking,zhu2024oftsr}.

To alleviate these issues, structured state space models (SSMs) have emerged as a compelling alternative due to their linear computational complexity and ability to capture long-range dependencies~\cite{ref12,ref13,ref14,wu2025sscm}. Among them, Mamba~\cite{ref15}, a selective SSM with hardware-friendly design, has demonstrated outstanding performance in natural language processing and has gained increasing attention in the computer vision community~\cite{ref16,ref17,wang2025angio,wang2022dense,wang2024cardiovascular}. Unlike standard transformers, Mamba enables efficient sequence-to-sequence modeling by dynamically learning input-dependent transition dynamics, making it particularly suitable for high-quality image restoration tasks.

Recent studies have integrated Mamba into vision networks for image restoration. For example, MambaIR~\cite{ref18} first applied Mamba to image deraining, demonstrating its efficiency in spatial dependency modeling. Subsequently, several works have focused on improving its scanning mechanism~\cite{ref19,ref50,ref21,ref22,di2025qmambabsr}, such as adopting multi-directional or adaptive strategies to capture richer contextual information. These refinements further enhance feature interaction and contribute to better restoration performance.

However, directly applying standard Mamba to image deraining has limitations: it lacks frequency-domain awareness and sensitivity to fine details. Since rain streaks mainly occupy mid- and high-frequency bands, ignoring frequency information can lead to incomplete removal and residual artifacts. Additionally, Mamba’s sequential modeling struggles to capture subtle textures and structures, reducing its effectiveness for deraining tasks.

The contribution can be summarized as follows:
\begin{itemize}
    \item {We propose a frequency and gradient-enhanced state space model, termed DeRainMamba, which is built upon a U-Net architecture. By incorporating frequency-domain representations and gradient-level cues into the state space modeling process, DeRainMamba effectively captures both global spectral characteristics and local structural details, enabling high-quality image deraining.}
    \item {We propose a novel Frequency-Aware State Space Model (FASSM), which explicitly leverages the high-frequency characteristics of rain streaks in the frequency domain. By integrating frequency-domain priors with the global modeling capability of SSMs, FASSM effectively enhances the discrimination and removal of rain streaks.}
    \item {We introduce a Multi-direction Perception Convolution (MDPConv) to reinforce gradient-level cues for background texture reconstruction. This block employs detail-aware convolutional operators to recover fine-grained image structures, thereby improving the fidelity of background restoration after rain removal.}
\end{itemize}

\section{Method}
\subsection{Frequency-Aware State Space Model}
Transformer-based image restoration networks have achieved notable success using attention and feed-forward modules. However, their high computational cost limits practical deployment. Recently, State Space Models (SSMs) have emerged as a more efficient alternative, providing a better balance between global context modeling and computational efficiency for image deraining. Despite this, existing SSM-based methods struggle with complex, overlapping rain streaks. Notably, Rain degradation mainly affects the mid- and high-frequency components, displaying directional patterns in the frequency domain. This makes frequency-domain processing a promising approach for isolating rain streaks from background content, enabling more accurate removal and detail recovery. To this end, we propose the Frequency-Aware State Space Models (FASSM) — a novel frequency-domain module that integrates SSMs to enhance rain streak perception and removal:
\begin{equation}
\begin{aligned}
F_{\text{out}} = \text{VSSM}(\text{LN}(F_{in})) + \text{RFM}(F_{in}) + s \cdot F_{in} ,
\end{aligned}
\label{FFT1}
\end{equation}

Given an input $F_{\text{in}}$,  we first apply Layer Normalization (LN) to normalize the feature distribution.
The normalized feature is then fed into two parallel branches: the Vision State Space Module (VSSM) for spatial modeling, and the Residual Fourier Module (RFM) for frequency-domain enhancement. A learnable scalar $s$ adaptively fuses the original and refined features. This enables effective integration of spatial and frequency information for improved rain removal.

\begin{figure}[!t]
    \centering
    \includegraphics[width=3.5in]{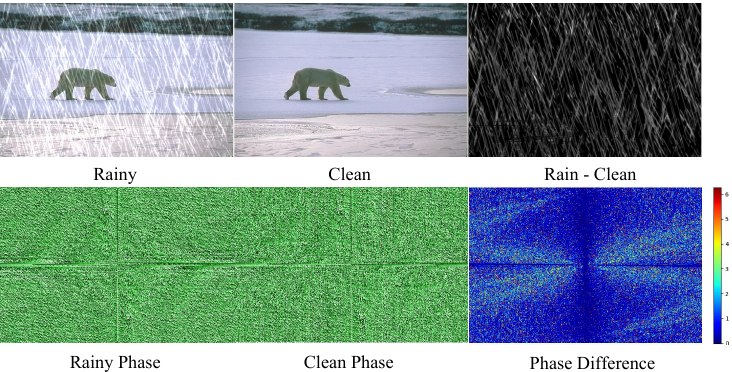}  
    \caption{Visualization of phase differences between rain-free and rainy images. }
    \label{fig:phase_diff}
\end{figure}

In the proposed RFM, the input \(F_{in}\) is transformed to the frequency domain by FFT, separating amplitude and phase for targeted processing. In contrast to existing approaches~\cite{ref45,ref47,ref23,ref20}, our observations indicate that rain streaks introduce significant distortions to both the amplitude and phase spectra of images. As shown in Fig.~\ref{fig:phase_diff}, the phase spectra of rainy and clean images exhibit notable differences, particularly in high-frequency regions where structural details such as edges and contours are concentrated. Ignoring these phase distortions may result in visual artifacts, including ghosting, edge blurring, and structural misalignment. To address this, RFM simultaneously refines amplitude and phase through parallel convolutional branches, enabling more accurate frequency-domain restoration.

\begin{equation}
\begin{aligned}
\mathcal{A}(X) &= |\mathcal{F}(X)|, \\
\mathcal{P}(X) &= \angle \mathcal{F}(X), 
\end{aligned}
\label{FFT2}
\end{equation}
where $\mathcal{F}(X)$ denotes the 2D Fourier Transform of the input $X$, and $\mathcal{A}(X)$, $\mathcal{P}(X)$ represent the amplitude and phase spectra, respectively. As rain streaks mainly affect high-frequency amplitude, a two-layer $1\times1$ convolution with ReLU is applied to enhance details and suppress noise. For phase distortions, a lightweight convolution is used to preserve structural information. The refined amplitude and phase are combined and transformed back to the spatial domain via IFFT. A $3\times3$ convolution fuses the restored features, followed by a residual depthwise convolution to further retain high-frequency details. The RFM can be represented as:
\begin{equation}
\begin{aligned}
\hat{X} &= \text{ReLU}( \text{ Conv}_{1 \times 1}(\mathcal{A}(X)) + \text{ Conv}_{1 \times 1}(\mathcal{P}(X))), \\
X_{out} &= \text{ Conv}_{1 \times 1}(\mathcal{F}^{-1}(\hat{X})) + \text{DWConv}(X) ,
\end{aligned}
\label{FFT3}
\end{equation}
where \text{DWConv} represents a $3 \times 3$ depthwise separable convolution that enhances high-frequency detail features, and $\mathcal{F}^{-1}$ denotes the inverse Fourier Transform (IFFT).
\begin{figure*}[!t]
    \centering
    \includegraphics[width=\textwidth,height=8cm,keepaspectratio]{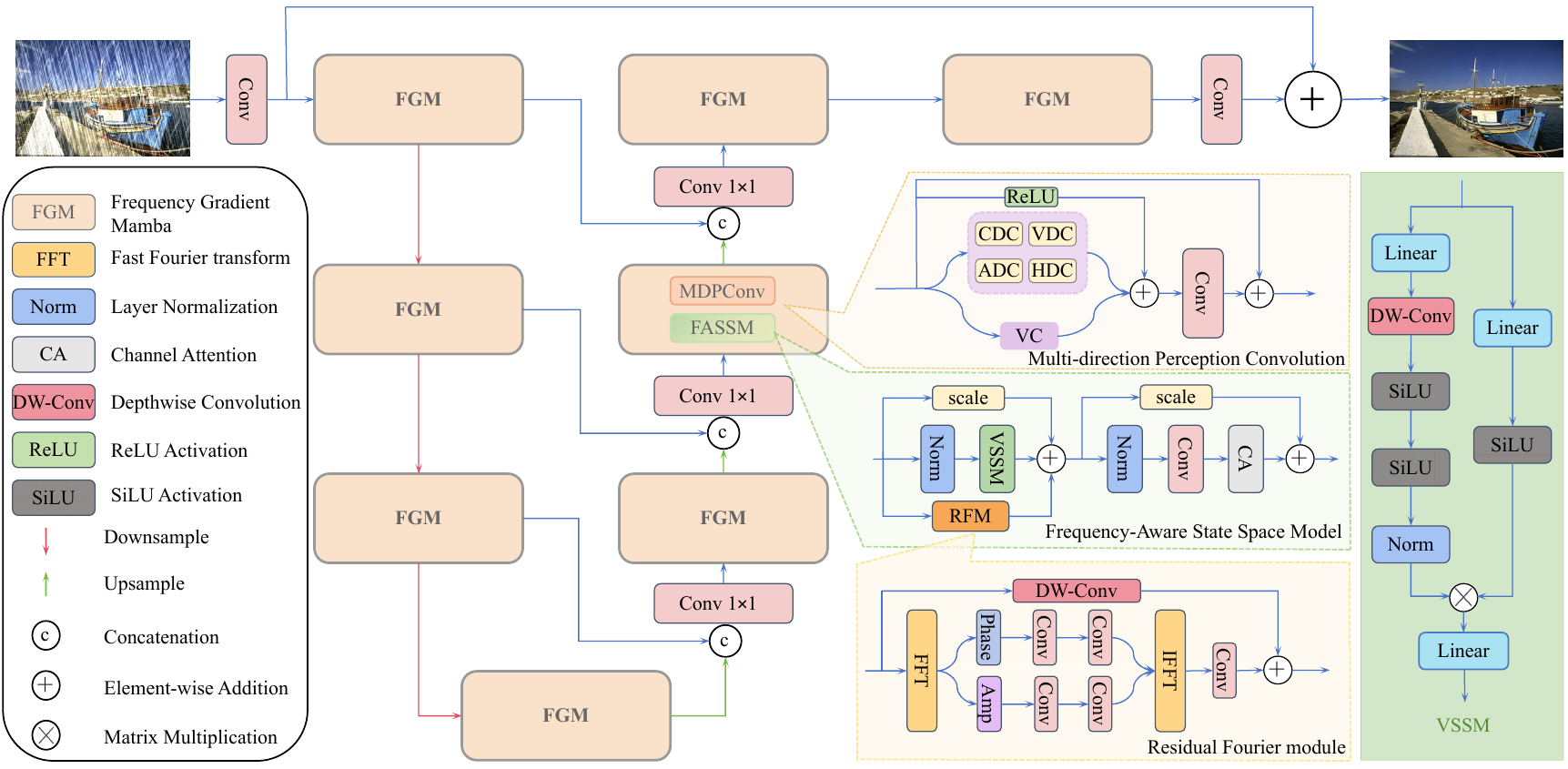}    
    \caption{Overall architecture of the proposed DeRainMamba. The network consists of multiple Frequency Gradient Mamba (FGM) blocks, each comprising a Frequency-Aware State Space Module (FASSM) for global modeling and an MDPConv branch for gradient-based detail enhancement. The two branches are fused to jointly suppress rain streaks and restore image details.}
    \label{fig:model_architecture}
\end{figure*}

\subsection{Multi-direction Perception Convolution}
To enhance background detail restoration in derained images, we propose a Multi-direction Perception Convolution (MDPConv), as shown in Fig.~\ref{fig:model_architecture}, based on Differential Convolution (DEConv)~\cite{ref24}. This module enhances the representational capacity of convolutional layers by explicitly incorporating gradient information, particularly for recovering high-frequency details such as edges and contours. MDPConv captures fine-grained structural cues by computing pixel-wise differences in both horizontal and vertical directions to extract gradient features. Unlike conventional convolution methods, MDPConv first derives gradient maps through differential operations and then applies convolutional kernels to process these features. This design enables the network to better distinguish between rain streaks and background details, thereby preserving intrinsic image structures while effectively removing rain artifacts. Specifically, we integrate five types of differential convolutions, including Horizontal Differential Convolution (HDC), vanilla convolution (VC), Central Difference Convolution (CDC), Angular Difference Convolution (ADC), and Vertical Differential Convolution (VDC), to extract more comprehensive structural information. To optimize computational efficiency, all differential convolutions are unified into a single convolutional operation. Through a re-parameterization strategy, the outputs of the five differential branches are merged into an equivalent convolution kernel, eliminating the overhead caused by parallel convolutional layers.
\begin{equation}
\begin{aligned}
F_{\text{out}} &= \text{MDPConv}(F_{\text{in}}) = \sum_{i=1}^{5} F_{\text{in}} * K_i \\
&= F_{\text{in}} * \left( \sum_{i=1}^{5} K_i \right) = F_{\text{in}} * K_{\text{eq}},
\end{aligned}
\label{gradient1}
\end{equation}
where \(F_{\text{in}}\) and \(F_{\text{out}}\) denote the input and output feature maps, respectively. 
\(K_i\) represents the \(i\)-th differential convolution kernel (HDC, VC, CDC, ADC, and VDC), and \(*\) denotes the convolution operation. To reduce computational overhead, we merge the five kernels \(\{K_i\}_{i=1}^{5}\) into a single equivalent kernel \(K_{\text{eq}}\) via re-parameterization.

Importantly, MDPConv operates synergistically with RFM: while RFM refines amplitude and phase components in the frequency domain to suppress rain streaks, MDPConv leverages gradient cues in the spatial domain to recover structural details. This complementary design enables joint global spectral restoration and local detail preservation, thereby enhancing overall deraining performance. The overall architecture of our proposed network is depicted in Fig.~\ref{fig:model_architecture}.

\subsection{Loss Function}
Following previous works \cite{ref5,ref7}, we utilize the common L1 loss \(\mathcal{L}_{1}\) for training. Furthermore, to enhance the network's ability to capture details, we also employ the frequency loss \(\mathcal{L}_{Freq}\) \cite{ref25,ref26}. The total loss is presented as follows:
\begin{equation}
\label{loss}
\begin{aligned}
\mathcal{L}_{total} &= \lambda_{1} \mathcal{L}_{1}(\mathcal{B}, \mathcal{B}_{gt}) + \lambda_{2} \mathcal{L}_{Freq}(\mathcal{B}, \mathcal{B}_{gt}) ,
\end{aligned}
\end{equation}
where $\mathcal{B}$ and $\mathcal{B}_{gt}$ denote the predicted output and the corresponding ground truth, respectively. The parameters $\lambda_{1}$ and $\lambda_{2}$ are balancing factors. In our experiments, we set $\lambda_{1}$ and $\lambda_{2}$ to 1, 0.1, respectively.

\setlength{\tabcolsep}{3pt}
\begin{table}[!t]
\centering
\caption{Quantitative PSNR $\uparrow$ and SSIM comparisons with existing state-of-the-art image deraining methods. }
\label{tab:quantitative_results}
\resizebox{\columnwidth}{!}{%
\begin{tabular}{l|cc|cc|cc|cc}
\toprule
\multicolumn{1}{c|}{\multirow{2}{*}{\textbf{Method}}} & \multicolumn{2}{c|}{\textbf{Rain200L}} & \multicolumn{2}{c|}{\textbf{Rain200H}} & \multicolumn{2}{c|}{\textbf{DID-Data}} & \multicolumn{2}{c}{\textbf{DDN-Data}} \\
 & PSNR $\uparrow$ & SSIM $\uparrow$ & PSNR $\uparrow$ & SSIM $\uparrow$ & PSNR $\uparrow$ & SSIM  & PSNR $\uparrow$ & SSIM  \\
\midrule
PReNet   & 37.80 & 0.9814 & 29.04 & 0.8991 & 33.17 & 0.9481 & 32.60 & 0.9459 \\
MSPFN    & 38.58 & 0.9827 & 29.36 & 0.9034 & 33.72 & 0.9550 & 32.99 & 0.9333 \\
RCDNet   & 39.17 & 0.9885 & 30.24 & 0.9048 & 34.08 & 0.9532 & 33.04 & 0.9472 \\
MPRNet   & 39.47 & 0.9825 & 30.67 & 0.9110 & 33.99 & 0.9590 & 33.10 & 0.9347 \\
DualGCN  & 40.73 & 0.9886 & 31.15 & 0.9125 & 34.37 & 0.9620 & 33.01 & 0.9489 \\
SPDNet   & 40.50 & 0.9875 & 31.28 & 0.9207 & 34.57 & 0.9560 & 33.15 & 0.9457 \\
Uformer  & 40.20 & 0.9860 & 30.80 & 0.9105 & 35.02 & 0.9621 & 33.95 & 0.9545 \\

Restormer& 40.99 & 0.9890 & 32.00 & 0.9329 & 35.29 & 0.9641 & 34.20 & 0.9571 \\
IDT      & 40.74 & 0.9884 & 32.10 & \underline{0.9344} & 34.89 & 0.9623 & 33.84 & 0.9549 \\
DLINet   & 40.91 & 0.9231 & -     & -      & 33.61 & 0.9514 & 33.61 & 0.9514 \\
DRSformer& \underline{41.23} & 0.9894 & 32.17 & 0.9326 & \underline{35.35} & \textbf{0.9646} & \underline{34.35} & \textbf{0.9588} \\
MambaIR  & 41.13 & \underline{0.9895} & \underline{32.18} & 0.9295 & 35.05 & 0.9612 & 34.00 & 0.9554 \\
TAMambaIR & 41.25 & 0.9896 & 32.19 & 0.9345 & -     & -    & -     & -     \\
\textbf{Ours} & \textbf{41.71} & \textbf{0.9899} & \textbf{32.63} & \textbf{0.9397} & \textbf{35.52} &\underline{0.9642} & \textbf{34.43} & \underline{0.9581} \\
\bottomrule
\end{tabular}%
}
\end{table}

\begin{table}[!t]
\centering
\caption{Comparisons of model complexity against state-of-the-art methods. The input size is $256 \times 256$ pixels, “Params (M)” denotes the number of parameters in millions.}
\label{tab:params}
\begin{threeparttable}
\begin{tabular}{c|cccccc}
\toprule
\textbf{Methods} & \makecell{Uformer} & \makecell{Restormer} & \makecell{DRSformer} & \makecell{MambaIR} & \textbf{Ours} \\
\midrule
\makecell{Params (M) $\downarrow$} 
& 50.880 & 26.127 & 33.660 & 31.506 & \textbf{27.797} \\
\bottomrule
\end{tabular}
\end{threeparttable}
\end{table}

\section{Experiments}
\subsection{Experimental Settings}
\textbf{Datasets.} Our method is evaluated on several widely used deraining benchmarks. Rain200H and Rain200L~\cite{ref28} simulate heavy and light rain conditions, respectively, providing a controlled environment for performance comparison. DID-Data~\cite{ref1} includes a wide range of rain streak densities and directions, facilitating the evaluation of adaptability to diverse rain patterns. DDN-Data~\cite{ref4} further extends the diversity of rain types, enabling a comprehensive assessment of model generalization.

\noindent{\textbf{Evaluation Metrics.} We use PSNR~\cite{ref31} and SSIM~\cite{ref2} as evaluation metrics. These two metrics are calculated on the luminance channel (Y channel) in the YCbCr color space. Higher PSNR and SSIM values indicate better image recovery quality, suggesting that the model performs well in enhancing image clarity and details.

\noindent{\textbf{Implementation Details.} Following prior research~\cite{ref5}, our DeRainMamba employs a 4-level encoder-decoder architecture. The model includes 4 refinement blocks and is trained with the AdamW optimizer using $\beta_1{=}0.9$, $\beta_2{=}0.999$, a weight decay of $1\times10^{-4}$, and a loss balance coefficient $\lambda_f{=}0.01$. The learning rate starts at $3\times10^{-4}$ and decays to $1\times10^{-6}$ via cosine scheduling~\cite{ref33} over 300K iterations. All experiments are conducted using PyTorch on an NVIDIA RTX 4090 GPU.

\noindent{\textbf{Comparisons with State-of-the-art Methods.} We compare our method against thirteen state-of-the-art methods, including: PReNet~\cite{ref2}, MSPFN~\cite{ref35}, RCDNet~\cite{ref36}, MPRNet~\cite{ref37}, DualGCN~\cite{ref38}, SPDNet~\cite{ref39}, Uformer~\cite{ref42}, Restormer~\cite{ref5}, IDT~\cite{ref40}, DLINet~\cite{ref41}, DRSformer~\cite{ref7}, MambaIR~\cite{ref18} and TAMambaIR~\cite{ref22}. } 
\begin{figure}[t]
  \centering
  \includegraphics[width=\linewidth]{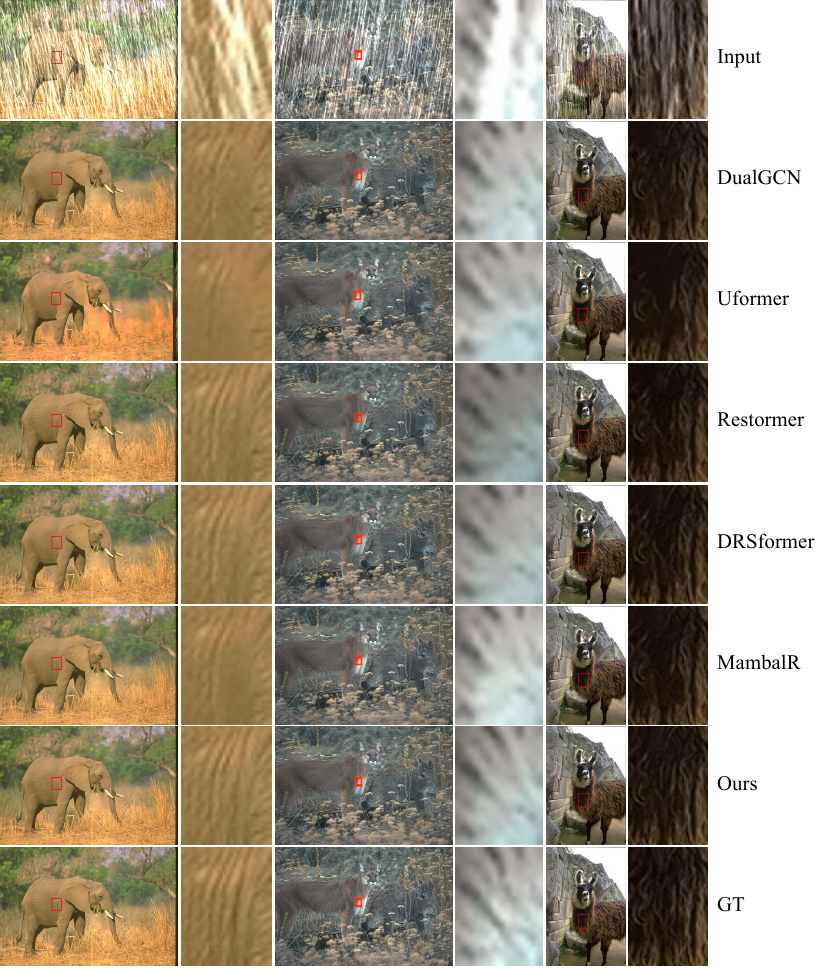}
  \caption{Visual comparison on the Rain200H dataset.}
  \label{fig:rain200h_visual}
\end{figure}

\subsection{Quantitative and Qualitative Results}
We compare our method with recent SOTA approaches, including DRSformer and MambaIR, on four standard datasets using PSNR and SSIM metrics. As shown in Table~\ref{tab:quantitative_results}, Our method achieves competitive or superior performance compared with recent methods. On Rain200H, it outperforms DRSformer by 0.45 dB in PSNR and 0.0068 in SSIM. Compared to MambaIR, our method achieves 0.44 dB and 0.0099 improvements, respectively. Similar gains are observed on DID-Data and DDN-Data, demonstrating the robustness of our approach. As show in Fig.~\ref{fig:rain200h_visual}, Qualitative results also show that our method removes rain streaks more effectively while preserving fine details.

\subsection{Comparison of model complexity} 
Model complexity is critical for real-world deployment. As shown in Table~\ref{tab:params}, our model has only 27.80M parameters, significantly fewer than Uformer (50.88M), DRSformer (33.66M), and MambaIR (31.51M). Notably, our model reduces 3.71M parameters compared to MambaIR while achieving superior deraining performance, highlighting its efficiency and suitability for resource-constrained devices.

\subsection{Ablation Study}
\begin{table}[t]
  \centering
  \caption{Ablation study on the components of DeRainMamba.}
  \setlength{\tabcolsep}{10pt} 
  \renewcommand{\arraystretch}{1.2} 
  \begin{tabular}{c c c | c c }
    \toprule
    \textbf{Baseline} & \textbf{RFM} & \textbf{MDPConv} & \textbf{PSNR}$\uparrow$ & \textbf{SSIM}$\uparrow$ \\
    \midrule
    \checkmark &             &             & 41.13 & 0.9904 \\
    \checkmark & \checkmark  &             & 41.55 & 0.9897 \\
    \checkmark &             & \checkmark  & 41.49 & 0.9894 \\
    \checkmark & \checkmark  & \checkmark  & \textbf{41.71} & \textbf{0.9899} \\
    \bottomrule
  \end{tabular}
  \label{tab:ablation_mamba}
\end{table}
To evaluate the effectiveness of each component in our framework, we conduct ablation studies on the baseline model by incrementally adding the Residual Fourier Module (RFM) and the proposed MDPConv. As shown in the table~\ref{tab:ablation_mamba}, introducing RFM alone improves the PSNR from 41.13 dB to 41.55 dB, demonstrating its ability to enhance detail restoration. Similarly, integrating only MDPConv yields a PSNR of 41.49 dB, validating the benefit of multi-directional gradient perception. When both RFM and MDPConv are incorporated, our model achieves the best performance with a PSNR of 41.71 dB and an SSIM of 0.9899, confirming the complementary roles of the two modules.

\section{Conclusion}
In this paper, we propose DeRainMamba, a lightweight yet effective image deraining method that integrates a Frequency-Aware State Space Module (FASSM) and a Multi-direction Perception Convolution (MDPConv). By jointly leveraging frequency-domain priors and gradient-level information, DeRainMamba removes rain degradation while reconstructing fine details. A re-parameterization strategy in MDPConv further enhances inference efficiency without sacrificing performance. Extensive results validate the effectiveness of our approach, though challenges remain under extreme rain conditions, pointing to future work on improving generalization to more complex scenarios.

\bibliographystyle{IEEEtran}
\bibliography{refs}  

\begin{thebibliography}{10}
\providecommand{\url}[1]{#1}
\csname url@samestyle\endcsname
\providecommand{\newblock}{\relax}
\providecommand{\bibinfo}[2]{#2}
\providecommand{\BIBentrySTDinterwordspacing}{\spaceskip=0pt\relax}
\providecommand{\BIBentryALTinterwordstretchfactor}{4}
\providecommand{\BIBentryALTinterwordspacing}{\spaceskip=\fontdimen2\font plus
\BIBentryALTinterwordstretchfactor\fontdimen3\font minus \fontdimen4\font\relax}
\providecommand{\BIBforeignlanguage}[2]{{%
\expandafter\ifx\csname l@#1\endcsname\relax
\typeout{** WARNING: IEEEtran.bst: No hyphenation pattern has been}%
\typeout{** loaded for the language `#1'. Using the pattern for}%
\typeout{** the default language instead.}%
\else
\language=\csname l@#1\endcsname
\fi
#2}}
\providecommand{\BIBdecl}{\relax}
\BIBdecl

\bibitem{gan2025enhancing}
L.~Gan, J.~Zhang, L.~Qu, Y.~Wang, S.~Wu, and X.~Sun, ``Enhancing zero-shot brain tumor subtype classification via fine-grained patch-text alignment,'' \emph{arXiv preprint arXiv:2508.01602}, 2025.

\bibitem{gan2025semamil}
L.~Gan, X.~Wu, J.~Zhang, Z.~Wang, L.~Qu, S.~Wu, and X.~Sun, ``Semamil: Semantic reordering with retrieval-guided state space modeling for whole slide image classification,'' \emph{arXiv preprint arXiv:2509.00442}, 2025.

\bibitem{li2023ntire}
Y.~Li, Y.~Zhang, R.~Timofte, L.~Van~Gool, L.~Yu, Y.~Li, X.~Li, T.~Jiang, Q.~Wu, M.~Han \emph{et~al.}, ``Ntire 2023 challenge on efficient super-resolution: Methods and results,'' in \emph{Proceedings of the IEEE/CVF Conference on Computer Vision and Pattern Recognition}, 2023, pp. 1922--1960.

\bibitem{ren2024ninth}
B.~Ren, Y.~Li, N.~Mehta, R.~Timofte, H.~Yu, C.~Wan, Y.~Hong, B.~Han, Z.~Wu, Y.~Zou \emph{et~al.}, ``The ninth ntire 2024 efficient super-resolution challenge report,'' in \emph{Proceedings of the IEEE/CVF Conference on Computer Vision and Pattern Recognition}, 2024, pp. 6595--6631.

\bibitem{wang2025ntire}
Y.~Wang, Z.~Liang, F.~Zhang, L.~Tian, L.~Wang, J.~Li, J.~Yang, R.~Timofte, Y.~Guo, K.~Jin \emph{et~al.}, ``Ntire 2025 challenge on light field image super-resolution: Methods and results,'' in \emph{Proceedings of the Computer Vision and Pattern Recognition Conference}, 2025, pp. 1227--1246.

\bibitem{peng2020cumulative}
L.~Peng, A.~Jiang, Q.~Yi, and M.~Wang, ``Cumulative rain density sensing network for single image derain,'' \emph{IEEE Signal Processing Letters}, vol.~27, pp. 406--410, 2020.

\bibitem{wang2023decoupling}
Y.~Wang, L.~Peng, L.~Li, Y.~Cao, and Z.-J. Zha, ``Decoupling-and-aggregating for image exposure correction,'' in \emph{Proceedings of the IEEE/CVF conference on computer vision and pattern recognition}, 2023, pp. 18\,115--18\,124.

\bibitem{peng2024lightweight}
L.~Peng, Y.~Cao, Y.~Sun, and Y.~Wang, ``Lightweight adaptive feature de-drifting for compressed image classification,'' \emph{IEEE Transactions on Multimedia}, vol.~26, pp. 6424--6436, 2024.

\bibitem{peng2024towards}
L.~Peng, W.~Li, R.~Pei, J.~Ren, J.~Xu, Y.~Wang, Y.~Cao, and Z.-J. Zha, ``Towards realistic data generation for real-world super-resolution,'' \emph{arXiv preprint arXiv:2406.07255}, 2024.

\bibitem{wang2023brightness}
H.~Wang, L.~Peng, Y.~Sun, Z.~Wan, Y.~Wang, and Y.~Cao, ``Brightness perceiving for recursive low-light image enhancement,'' \emph{IEEE Transactions on Artificial Intelligence}, vol.~5, no.~6, pp. 3034--3045, 2023.

\bibitem{peng2021ensemble}
L.~Peng, A.~Jiang, H.~Wei, B.~Liu, and M.~Wang, ``Ensemble single image deraining network via progressive structural boosting constraints,'' \emph{Signal Processing: Image Communication}, vol.~99, p. 116460, 2021.

\bibitem{ren2024ultrapixel}
J.~Ren, W.~Li, H.~Chen, R.~Pei, B.~Shao, Y.~Guo, L.~Peng, F.~Song, and L.~Zhu, ``Ultrapixel: Advancing ultra high-resolution image synthesis to new peaks,'' \emph{Advances in Neural Information Processing Systems}, vol.~37, pp. 111\,131--111\,171, 2024.

\bibitem{yan2025textual}
Q.~Yan, A.~Jiang, K.~Chen, L.~Peng, Q.~Yi, and C.~Zhang, ``Textual prompt guided image restoration,'' \emph{Engineering Applications of Artificial Intelligence}, vol. 155, p. 110981, 2025.

\bibitem{peng2024efficient}
L.~Peng, Y.~Cao, R.~Pei, W.~Li, J.~Guo, X.~Fu, Y.~Wang, and Z.-J. Zha, ``Efficient real-world image super-resolution via adaptive directional gradient convolution,'' \emph{arXiv preprint arXiv:2405.07023}, 2024.

\bibitem{conde2024real}
M.~V. Conde, Z.~Lei, W.~Li, I.~Katsavounidis, R.~Timofte, M.~Yan, X.~Liu, Q.~Wang, X.~Ye, Z.~Du \emph{et~al.}, ``Real-time 4k super-resolution of compressed avif images. ais 2024 challenge survey,'' in \emph{Proceedings of the IEEE/CVF Conference on Computer Vision and Pattern Recognition}, 2024, pp. 5838--5856.

\bibitem{peng2025directing}
L.~Peng, X.~Di, Z.~Feng, W.~Li, R.~Pei, Y.~Wang, X.~Fu, Y.~Cao, and Z.-J. Zha, ``Directing mamba to complex textures: An efficient texture-aware state space model for image restoration,'' \emph{arXiv preprint arXiv:2501.16583}, 2025.

\bibitem{peng2025pixel}
L.~Peng, A.~Wu, W.~Li, P.~Xia, X.~Dai, X.~Zhang, X.~Di, H.~Sun, R.~Pei, Y.~Wang \emph{et~al.}, ``Pixel to gaussian: Ultra-fast continuous super-resolution with 2d gaussian modeling,'' \emph{arXiv preprint arXiv:2503.06617}, 2025.

\bibitem{peng2025boosting}
L.~Peng, Y.~Wang, X.~Di, X.~Fu, Y.~Cao, Z.-J. Zha \emph{et~al.}, ``Boosting image de-raining via central-surrounding synergistic convolution,'' in \emph{Proceedings of the AAAI Conference on Artificial Intelligence}, vol.~39, no.~6, 2025, pp. 6470--6478.

\bibitem{he2024latent}
Y.~He, L.~Peng, L.~Wang, and J.~Cheng, ``Latent degradation representation constraint for single image deraining,'' in \emph{ICASSP 2024-2024 IEEE International Conference on Acoustics, Speech and Signal Processing (ICASSP)}.\hskip 1em plus 0.5em minus 0.4em\relax IEEE, 2024, pp. 3155--3159.

\bibitem{peng2024unveiling}
L.~Peng, W.~Li, J.~Guo, X.~Di, H.~Sun, Y.~Li, R.~Pei, Y.~Wang, Y.~Cao, and Z.-J. Zha, ``Unveiling hidden details: A raw data-enhanced paradigm for real-world super-resolution,'' \emph{arXiv preprint arXiv:2411.10798}, 2024.

\bibitem{he2024dual}
Y.~He, A.~Jiang, L.~Jiang, L.~Peng, Z.~Wang, and L.~Wang, ``Dual-path coupled image deraining network via spatial-frequency interaction,'' in \emph{2024 IEEE International Conference on Image Processing (ICIP)}.\hskip 1em plus 0.5em minus 0.4em\relax IEEE, 2024, pp. 1452--1458.

\bibitem{he2024multi}
Y.~He, L.~Peng, Q.~Yi, C.~Wu, and L.~Wang, ``Multi-scale representation learning for image restoration with state-space model,'' \emph{arXiv preprint arXiv:2408.10145}, 2024.

\bibitem{pan2025enhance}
J.~Pan, Y.~Liu, X.~He, L.~Peng, J.~Li, Y.~Sun, and X.~Huang, ``Enhance then search: An augmentation-search strategy with foundation models for cross-domain few-shot object detection,'' in \emph{Proceedings of the Computer Vision and Pattern Recognition Conference}, 2025, pp. 1548--1556.

\bibitem{wu2025dropout}
C.~Wu, L.~Wang, L.~Peng, D.~Lu, and Z.~Zheng, ``Dropout the high-rate downsampling: A novel design paradigm for uhd image restoration,'' in \emph{2025 IEEE/CVF Winter Conference on Applications of Computer Vision (WACV)}.\hskip 1em plus 0.5em minus 0.4em\relax IEEE, 2025, pp. 2390--2399.

\bibitem{jiang2024dalpsr}
A.~Jiang, Z.~Wei, L.~Peng, F.~Liu, W.~Li, and M.~Wang, ``Dalpsr: Leverage degradation-aligned language prompt for real-world image super-resolution,'' \emph{arXiv preprint arXiv:2406.16477}, 2024.

\bibitem{ignatov2025rgb}
A.~Ignatov, G.~Perevozchikov, R.~Timofte, W.~Pan, S.~Wang, D.~Zhang, Z.~Ran, X.~Li, S.~Ju, D.~Zhang \emph{et~al.}, ``Rgb photo enhancement on mobile gpus, mobile ai 2025 challenge: Report,'' in \emph{Proceedings of the Computer Vision and Pattern Recognition Conference}, 2025, pp. 1922--1933.

\bibitem{du2024fc3dnet}
Z.~Du, L.~Peng, Y.~Wang, Y.~Cao, and Z.-J. Zha, ``Fc3dnet: A fully connected encoder-decoder for efficient demoir{\'e}ing,'' in \emph{2024 IEEE International Conference on Image Processing (ICIP)}.\hskip 1em plus 0.5em minus 0.4em\relax IEEE, 2024, pp. 1642--1648.

\bibitem{jin2024mipi}
X.~Jin, C.~Guo, X.~Li, Z.~Yue, C.~Li, S.~Zhou, R.~Feng, Y.~Dai, P.~Yang, C.~C. Loy \emph{et~al.}, ``Mipi 2024 challenge on few-shot raw image denoising: Methods and results,'' in \emph{Proceedings of the IEEE/CVF Conference on Computer Vision and Pattern Recognition}, 2024, pp. 1153--1161.

\bibitem{sun2024beyond}
H.~Sun, W.~Li, J.~Liu, K.~Zhou, Y.~Chen, Y.~Guo, Y.~Li, R.~Pei, L.~Peng, and Y.~Yang, ``Beyond pixels: Text enhances generalization in real-world image restoration,'' \emph{arXiv preprint arXiv:2412.00878}, 2024.

\bibitem{qi2025data}
X.~Qi, R.~Li, L.~Peng, Q.~Ling, J.~Yu, Z.~Chen, P.~Chang, M.~Han, and J.~Xiao, ``Data-free knowledge distillation with diffusion models,'' \emph{arXiv preprint arXiv:2504.00870}, 2025.

\bibitem{feng2025pmq}
Z.~Feng, L.~Peng, X.~Di, Y.~Guo, W.~Li, Y.~Zhang, R.~Pei, Y.~Wang, Y.~Cao, and Z.-J. Zha, ``Pmq-ve: Progressive multi-frame quantization for video enhancement,'' \emph{arXiv preprint arXiv:2505.12266}, 2025.

\bibitem{xia2024s3mamba}
P.~Xia, L.~Peng, X.~Di, R.~Pei, Y.~Wang, Y.~Cao, and Z.-J. Zha, ``S3mamba: Arbitrary-scale super-resolution via scaleable state space model,'' \emph{arXiv preprint arXiv:2411.11906}, vol.~6, 2024.

\bibitem{pengboosting}
L.~Peng, W.~Li, J.~Guo, X.~Di, H.~Sun, Y.~Li, R.~Pei, Y.~Wang, Y.~Cao, and Z.-J. Zha, ``Boosting real-world super-resolution with raw data: a new perspective, dataset and baseline.''

\bibitem{suntext}
H.~Sun, W.~Li, J.~Liu, K.~Zhou, Y.~Chen, Y.~Guo, Y.~Li, R.~Pei, L.~Peng, and Y.~Yang, ``Text boosts generalization: A plug-and-play captioner for real-world image restoration.''

\bibitem{yakovenko2025aim}
A.~Yakovenko, G.~Chakvetadze, I.~Khrapov, M.~Zhelezov, D.~Vatolin, R.~Timofte, Y.~Oh, J.~Kwon, J.~Park, N.~I. Cho \emph{et~al.}, ``Aim 2025 low-light raw video denoising challenge: Dataset, methods and results,'' \emph{arXiv preprint arXiv:2508.16830}, 2025.

\bibitem{xu2025camel}
H.~Xu, L.~Peng, S.~Song, X.~Liu, M.~Jun, S.~Li, J.~Yu, and X.~Mao, ``Camel: Energy-aware llm inference on resource-constrained devices,'' \emph{arXiv preprint arXiv:2508.09173}, 2025.

\bibitem{wu2025robustgs}
A.~Wu, L.~Peng, X.~Di, X.~Dai, C.~Wu, Y.~Wang, X.~Fu, Y.~Cao, and Z.-J. Zha, ``Robustgs: Unified boosting of feedforward 3d gaussian splatting under low-quality conditions,'' \emph{arXiv preprint arXiv:2508.03077}, 2025.

\bibitem{li2023cross}
W.~Li, J.~Li, G.~Gao, W.~Deng, J.~Zhou, J.~Yang, and G.-J. Qi, ``Cross-receptive focused inference network for lightweight image super-resolution,'' \emph{IEEE Transactions on Multimedia}, vol.~26, pp. 864--877, 2023.

\bibitem{li2023survey}
W.~Li, M.~Wang, K.~Zhang, J.~Li, X.~Li, Y.~Zhang, G.~Gao, W.~Deng, and C.-W. Lin, ``Survey on deep face restoration: From non-blind to blind and beyond,'' \emph{arXiv preprint arXiv:2309.15490}, 2023.

\bibitem{li2024efficient1}
W.~Li, J.~Li, G.~Gao, W.~Deng, J.~Yang, G.-J. Qi, and C.-W. Lin, ``Efficient image super-resolution with feature interaction weighted hybrid network,'' \emph{IEEE Transactions on Multimedia}, 2024.

\bibitem{li2024efficient2}
W.~Li, H.~Guo, X.~Liu, K.~Liang, J.~Hu, Z.~Ma, and J.~Guo, ``Efficient face super-resolution via wavelet-based feature enhancement network,'' in \emph{Proceedings of the 32nd ACM International Conference on Multimedia}, 2024, pp. 4515--4523.

\bibitem{li2025fouriersr}
W.~Li, H.~Guo, Y.~Hou, and Z.~Ma, ``Fouriersr: A fourier token-based plugin for efficient image super-resolution,'' \emph{arXiv preprint arXiv:2503.10043}, 2025.

\bibitem{li2025dual}
W.~Li, H.~Guo, Y.~Hou, G.~Gao, and Z.~Ma, ``Dual-domain modulation network for lightweight image super-resolution,'' \emph{arXiv preprint arXiv:2503.10047}, 2025.

\bibitem{li2025self}
W.~Li, X.~Wang, H.~Guo, G.~Gao, and Z.~Ma, ``Self-supervised selective-guided diffusion model for old-photo face restoration,'' in \emph{NeurIPs}, 2025.

\bibitem{zheng2025towards}
Y.~Zheng, B.~Zhong, Q.~Liang, S.~Zhang, G.~Li, X.~Li, and R.~Ji, ``Towards universal modal tracking with online dense temporal token learning,'' \emph{IEEE Transactions on Pattern Analysis and Machine Intelligence}, 2025.

\bibitem{zheng2024odtrack}
Y.~Zheng, B.~Zhong, Q.~Liang, Z.~Mo, S.~Zhang, and X.~Li, ``Odtrack: Online dense temporal token learning for visual tracking,'' in \emph{Proceedings of the AAAI conference on artificial intelligence}, vol.~38, no.~7, 2024, pp. 7588--7596.

\bibitem{zheng2025decoupled}
Y.~Zheng, B.~Zhong, Q.~Liang, N.~Li, and S.~Song, ``Decoupled spatio-temporal consistency learning for self-supervised tracking,'' in \emph{Proceedings of the AAAI Conference on Artificial Intelligence}, vol.~39, no.~10, 2025, pp. 10\,635--10\,643.

\bibitem{zheng2023toward}
Y.~Zheng, B.~Zhong, Q.~Liang, G.~Li, R.~Ji, and X.~Li, ``Toward unified token learning for vision-language tracking,'' \emph{IEEE Transactions on Circuits and Systems for Video Technology}, vol.~34, no.~4, pp. 2125--2135, 2023.

\bibitem{zheng2022leveraging}
Y.~Zheng, B.~Zhong, Q.~Liang, Z.~Tang, R.~Ji, and X.~Li, ``Leveraging local and global cues for visual tracking via parallel interaction network,'' \emph{IEEE Transactions on Circuits and Systems for Video Technology}, vol.~33, no.~4, pp. 1671--1683, 2022.

\bibitem{duan2025dit4sr}
Z.-P. Duan, J.~Zhang, X.~Jin, Z.~Zhang, Z.~Xiong, D.~Zou, J.~Ren, C.-L. Guo, and C.~Li, ``Dit4sr: Taming diffusion transformer for real-world image super-resolution,'' in \emph{Proceedings of the IEEE/CVF International Conference on Computer Vision}, 2025.

\bibitem{chen2025mixnet}
W.~Chen, S.~Sun, Y.~Zhang, and Z.~Zheng, ``Mixnet: Efficient global modeling for ultra-high-definition image restoration,'' \emph{Neurocomputing}, p. 131130, 2025.

\bibitem{chen2024towards}
H.~Chen, X.~Chen, C.~Wu, Z.~Zheng, J.~Pan, and X.~Fu, ``Towards ultra-high-definition image deraining: A benchmark and an efficient method,'' \emph{arXiv preprint arXiv:2405.17074}, 2024.

\bibitem{wu2025adaptive}
C.~Wu, L.~Wang, X.~Su, and Z.~Zheng, ``Adaptive feature selection modulation network for efficient image super-resolution,'' \emph{IEEE Signal Processing Letters}, 2025.

\bibitem{wu2024rethinking}
C.~Wu, Z.~Zheng, P.~Dai, C.~Shan, and X.~Jia, ``Rethinking image deraining via text-guided detail reconstruction,'' in \emph{2024 IEEE International Conference on Multimedia and Expo (ICME)}.\hskip 1em plus 0.5em minus 0.4em\relax IEEE, 2024, pp. 1--6.

\bibitem{1}
Y.~Gong, Z.~Zhong, Y.~Qu, Z.~Luo, R.~Ji, and M.~Jiang, ``Cross-modality perturbation synergy attack for person re-identification,'' \emph{Advances in Neural Information Processing Systems}, vol.~37, pp. 23\,352--23\,377, 2024.

\bibitem{2}
Y.~Gong, L.~Huang, and L.~Chen, ``Person re-identification method based on color attack and joint defence,'' in \emph{CVPR, 2022}, 2022, pp. 4313--4322.

\bibitem{3}
------, ``Eliminate deviation with deviation for data augmentation and a general multi-modal data learning method,'' \emph{arXiv preprint arXiv:2101.08533}, 2021.

\bibitem{4}
Y.~Gong, C.~Zhang, Y.~Hou, L.~Chen, and M.~Jiang, ``Beyond dropout: Robust convolutional neural networks based on local feature masking,'' in \emph{2024 International Joint Conference on Neural Networks (IJCNN)}.\hskip 1em plus 0.5em minus 0.4em\relax IEEE, 2024, pp. 1--8.

\bibitem{5}
Y.~Gong, Y.~Hou, C.~Zhang, and M.~Jiang, ``Beyond augmentation: Empowering model robustness under extreme capture environments,'' in \emph{2024 International Joint Conference on Neural Networks (IJCNN)}.\hskip 1em plus 0.5em minus 0.4em\relax IEEE, 2024, pp. 1--8.

\bibitem{6}
Y.~Gong, Q.~Zeng, D.~Xu, Z.~Wang, and M.~Jiang, ``Cross-modality attack boosted by gradient-evolutionary multiform optimization,'' \emph{arXiv preprint arXiv:2409.17977}, 2024.

\bibitem{7}
Y.~Gong, J.~Li, L.~Chen, and M.~Jiang, ``Exploring color invariance through image-level ensemble learning,'' \emph{arXiv preprint arXiv:2401.10512}, 2022.

\bibitem{lu2024mace}
S.~Lu, Z.~Wang, L.~Li, Y.~Liu, and A.~W.-K. Kong, ``Mace: Mass concept erasure in diffusion models,'' in \emph{Proceedings of the IEEE/CVF Conference on Computer Vision and Pattern Recognition}, 2024, pp. 6430--6440.

\bibitem{lu2023tf}
S.~Lu, Y.~Liu, and A.~W.-K. Kong, ``Tf-icon: Diffusion-based training-free cross-domain image composition,'' in \emph{Proceedings of the IEEE/CVF International Conference on Computer Vision}, 2023, pp. 2294--2305.

\bibitem{lu2024robust}
S.~Lu, Z.~Zhou, J.~Lu, Y.~Zhu, and A.~W.-K. Kong, ``Robust watermarking using generative priors against image editing: From benchmarking to advances,'' \emph{arXiv preprint arXiv:2410.18775}, 2024.

\bibitem{lu2025does}
S.~Lu, Z.~Lian, Z.~Zhou, S.~Zhang, C.~Zhao, and A.~W.-K. Kong, ``Does flux already know how to perform physically plausible image composition?'' \emph{arXiv preprint arXiv:2509.21278}, 2025.

\bibitem{zhou2025dragflow}
Z.~Zhou, S.~Lu, S.~Leng, S.~Zhang, Z.~Lian, X.~Yu, and A.~W.-K. Kong, ``Dragflow: Unleashing dit priors with region based supervision for drag editing,'' \emph{arXiv preprint arXiv:2510.02253}, 2025.

\bibitem{li2025set}
L.~Li, S.~Lu, Y.~Ren, and A.~W.-K. Kong, ``Set you straight: Auto-steering denoising trajectories to sidestep unwanted concepts,'' \emph{arXiv preprint arXiv:2504.12782}, 2025.

\bibitem{gao2024eraseanything}
D.~Gao, S.~Lu, S.~Walters, W.~Zhou, J.~Chu, J.~Zhang, B.~Zhang, M.~Jia, J.~Zhao, Z.~Fan \emph{et~al.}, ``Eraseanything: Enabling concept erasure in rectified flow transformers,'' \emph{arXiv preprint arXiv:2412.20413}, 2024.

\bibitem{ren2025all}
Y.~Ren, S.~Lu, and A.~W.-K. Kong, ``All that glitters is not gold: Key-secured 3d secrets within 3d gaussian splatting,'' \emph{arXiv preprint arXiv:2503.07191}, 2025.

\bibitem{lu2022copy}
S.~Lu, X.~Hu, C.~Wang, L.~Chen, S.~Han, and Y.~Han, ``Copy-move image forgery detection based on evolving circular domains coverage,'' \emph{Multimedia Tools and Applications}, vol.~81, no.~26, pp. 37\,847--37\,872, 2022.

\bibitem{yang2025temporal}
S.~Yang, S.~Lu, S.~Wang, M.~H. Er, Z.~Zheng, and A.~C. Kot, ``Temporal-guided spiking neural networks for event-based human action recognition,'' \emph{arXiv preprint arXiv:2503.17132}, 2025.

\bibitem{yu2025visual}
X.~Yu, Z.~Chen, Y.~Zhang, S.~Lu, R.~Shen, J.~Zhang, X.~Hu, Y.~Fu, and S.~Yan, ``Visual document understanding and question answering: A multi-agent collaboration framework with test-time scaling,'' \emph{arXiv preprint arXiv:2508.03404}, 2025.

\bibitem{gao2025revoking}
D.~Gao, N.~Jiang, A.~Zhang, S.~Lu, Y.~Tang, W.~Zhou, W.~Zhang, and Z.~Fan, ``Revoking amnesia: Rl-based trajectory optimization to resurrect erased concepts in diffusion models,'' 2025.

\bibitem{zhu2024oftsr}
Y.~Zhu, R.~Wang, S.~Lu, J.~Li, H.~Yan, and K.~Zhang, ``Oftsr: One-step flow for image super-resolution with tunable fidelity-realism trade-offs,'' \emph{arXiv preprint arXiv:2412.09465}, 2024.

\bibitem{wu2025sscm}
X.~Wu, L.~Gan, S.~Wu, J.~Zhang, Y.~Ou, and X.~Sun, ``Sscm: A spatial-semantic consistent model for multi-contrast mri super-resolution,'' \emph{arXiv preprint arXiv:2509.18593}, 2025.

\bibitem{wang2025angio}
Z.~Wang, R.~Yi, X.~Wen, C.~Zhu, K.~Xu, and K.~He, ``Angio-diff: learning a self-supervised adversarial diffusion model for angiographic geometry generation: Z. wang et al.'' \emph{The Visual Computer}, pp. 1--13, 2025.

\bibitem{wang2022dense}
Z.~Wang and A.~Jiang, ``A dense prediction vit network for single image bokeh rendering,'' in \emph{Chinese Conference on Pattern Recognition and Computer Vision (PRCV)}.\hskip 1em plus 0.5em minus 0.4em\relax Springer, 2022, pp. 213--222.

\bibitem{wang2024cardiovascular}
Z.~Wang, R.~Yi, X.~Wen, C.~Zhu, and K.~Xu, ``Cardiovascular medical image and analysis based on 3d vision: A comprehensive survey,'' \emph{Meta-Radiology}, vol.~2, no.~4, p. 100102, 2024.

\bibitem{di2025qmambabsr}
X.~Di, L.~Peng, P.~Xia, W.~Li, R.~Pei, Y.~Cao, Y.~Wang, and Z.-J. Zha, ``Qmambabsr: Burst image super-resolution with query state space model,'' in \emph{Proceedings of the Computer Vision and Pattern Recognition Conference}, 2025, pp. 23\,080--23\,090.

\end{thebibliography}


\begin{thebibliography}{99}

\bibitem{ref1}
H. Zhang and V. M. Patel, "Density-Aware Single Image De-raining Using a Multi-stream Dense Network," in \textit{Proc. IEEE/CVF Conf. Comput. Vis. Pattern Recognit. (CVPR)}, Salt Lake City, UT, USA, 2018, pp. 695--704, doi: 10.1109/CVPR.2018.00079.

\bibitem{ref2}
D. Ren, W. Zuo, Q. Hu, P. Zhu and D. Meng, "Progressive Image Deraining Networks: A Better and Simpler Baseline," in \textit{Proc. IEEE/CVF Conf. Comput. Vis. Pattern Recognit. (CVPR)}, Long Beach, CA, USA, 2019, pp. 3932--3941, doi: 10.1109/CVPR.2019.00406.

\bibitem{ref3}
X. Fu, J. Huang, D. Zeng, Y. Huang, X. Ding and J. Paisley, "Removing Rain from Single Images via a Deep Detail Network," in \textit{Proc. IEEE Conf. Comput. Vis. Pattern Recognit. (CVPR)}, Honolulu, HI, USA, 2017, pp. 1715--1723, doi: 10.1109/CVPR.2017.186.

\bibitem{ref8}
L. Peng \textit{et al.}, “Boosting Image De-Raining via Central-Surrounding Synergistic Convolution”, in \textit{Proc. AAAI Conf. Artif. Intell. (AAAI)}, vol. 39, no. 6, pp. 6470--6478, Apr. 2025.

\bibitem{ref29}
L. Peng \textit{et al.}, "Efficient real-world image super-resolution via adaptive directional gradient convolution," arXiv:2405.07023, 2024.

\bibitem{ref4}
W. Yang, R. T. Tan, J. Feng, J. Liu, Z. Guo and S. Yan, "Deep Joint Rain Detection and Removal from a Single Image," in \textit{Proc. IEEE Conf. Comput. Vis. Pattern Recognit. (CVPR)}, Honolulu, HI, USA, 2017, pp. 1685--1694, doi: 10.1109/CVPR.2017.183.

\bibitem{ref44}
Z. Huang \textit{et al.}, "T2EA: Target-Aware Taylor Expansion Approximation Network for Infrared and Visible Image Fusion," in \textit{IEEE Trans. Circuits Syst. Video Technol.}, vol. 35, no. 5, pp. 4831--4845, May 2025, doi: 10.1109/TCSVT.2024.3524794.

\bibitem{ref5}
S. W. Zamir, A. Arora, S. Khan, M. Hayat, F. S. Khan and M. Yang, "Restormer: Efficient Transformer for High-Resolution Image Restoration," in \textit{Proc. IEEE/CVF Conf. Comput. Vis. Pattern Recognit. (CVPR)}, New Orleans, LA, USA, 2022, pp. 5718--5729, doi: 10.1109/CVPR52688.2022.00564.

\bibitem{ref6}
J. Liang, J. Cao, G. Sun, K. Zhang, L. Van Gool and R. Timofte, "SwinIR: Image Restoration Using Swin Transformer," in \textit{Proc. IEEE/CVF Int. Conf. Comput. Vis. Workshops (ICCVW)}, Montreal, BC, Canada, 2021, pp. 1833--1844, doi: 10.1109/ICCVW54120.2021.00210.

\bibitem{ref9}
S. Lei, Z. Shi and W. Mo, "Transformer-Based Multistage Enhancement for Remote Sensing Image Super-Resolution," in \textit{IEEE Trans. Geosci. Remote Sens.}, vol. 60, pp. 1--11, 2022, Art no. 5615611, doi: 10.1109/TGRS.2021.3136190.

\bibitem{ref7}
X. Chen, H. Li, M. Li and J. Pan, "Learning A Sparse Transformer Network for Effective Image Deraining," in \textit{Proc. IEEE/CVF Conf. Comput. Vis. Pattern Recognit. (CVPR)}, Vancouver, BC, Canada, 2023, pp. 5896--5905, doi: 10.1109/CVPR52729.2023.00571.

\bibitem{ref10}
R. Qian, X. Dong, P. Zhang, Y. Zang, S. Ding, D. Lin, and J. Wang, "Streaming long video understanding with large language models," in \textit{Adv. Neural Inf. Process. Syst. (NeurIPS)}, vol. 37, 2024, pp. 119336--119360.

\bibitem{ref11}
Y. Tay, M. Dehghani, D. Bahri, and D. Metzler, "Efficient Transformers: A Survey," \textit{ACM Comput. Surv.}, vol. 55, no. 6, Art. no. 109, pp. 1--28, Jun. 2023, doi: 10.1145/3530811.

\bibitem{ref43}
Z. Huang, Hu W, Zhu Z, et al. "TMSF: Taylor expansion approximation network with multi-stage feature representation for optical flow estimation," \textit{Digit. Signal Process.}, 2025, 162: 105157. https://doi.org/10.1016/j.dsp.2025.105157

\bibitem{ref12}
A. Gu, K. Goel, and C. Ré, "Efficiently modeling long sequences with structured state spaces," arXiv:2111.00396, 2021.

\bibitem{ref13}
J. T. H. Smith, A. Warrington, and S. W. Linderman, "Simplified state space layers for sequence modeling," arXiv:2208.04933, 2022.

\bibitem{ref14}
R. E. Kalman, "A New Approach to Linear Filtering and Prediction Problems," \textit{J. Basic Eng.}, vol. 82, no. 1, pp. 35--45, Mar. 1960, doi: 10.1115/1.3662552.

\bibitem{ref15}
A. Gu and T. Dao, "Mamba: Linear-time sequence modeling with selective state spaces," arXiv:2312.00752, 2023.

\bibitem{ref16}
Y. Liu et al., "VMamba: Visual State Space Model," \textit{in Adv. Neural Inf. Process. Syst.}, vol. 37, pp. 103031–103063, 2024. 

\bibitem{ref17}
X. Liu, C. Zhang, and L. Zhang, "Vision Mamba: A comprehensive survey and taxonomy," arXiv:2405.04404, 2024.

\bibitem{ref18}
Guo, Hang, et al. "Mambair: A simple baseline for image restoration with state-space model." European conference on computer vision. Cham: Springer Nature Switzerland, 2024.

\bibitem{ref19}
Y. He \textit{et al.}, "Multi-scale representation learning for image restoration with state-space model," arXiv:2408.10145, 2024.

\bibitem{ref50}
P. Xia, L. Peng, Di X, et al. "$\text{S}^{3}$Mamba: Arbitrary-Scale Super-Resolution via Scaleable State Space Model," arXiv preprint arXiv:2411.11906, 2024.

\bibitem{ref21}
H. Guo et al., "MambaIRv2: Attentive State Space Restoration," 2025 IEEE/CVF Conference on Computer Vision and Pattern Recognition (CVPR), Nashville, TN, USA, 2025, pp. 28124-28133, doi: 10.1109/CVPR52734.2025.02619.

\bibitem{ref22}
L. Peng \textit{et al.}, "Directing Mamba to complex textures: An efficient texture-aware state space model for image restoration," arXiv:2501.16583, 2025.

\bibitem{ref20}
X. Luan \textit{et al.}, "FMambaIR: A Hybrid State-Space Model and Frequency Domain for Image Restoration," in \textit{IEEE Trans. Geosci. Remote Sens.}, vol. 63, pp. 1--14, 2025, Art no. 4201614, doi: 10.1109/TGRS.2025.3526927.

\bibitem{ref45}
Xiao. Yi \textit{et al.}, Z. Zou, H. Yu, J. Huang, and F. Zhao, "FreqMamba: Viewing Mamba from a Frequency Perspective for Image Deraining," in \textit{Proc. ACM Int. Conf. Multimedia (MM)}, 2024, pp. 1905--1914, doi: 10.1145/3664647.3680862.

\bibitem{ref47}
Li. Dong \textit{et al.}, "Fouriermamba: Fourier learning integration with state space models for image deraining." arXiv preprint arXiv:2405.19450 (2024).

\bibitem{ref23}
Y. Xiao, Q. Yuan, K. Jiang, Y. Chen, Q. Zhang and C. -W. Lin, "Frequency-Assisted Mamba for Remote Sensing Image Super-Resolution," in \textit{IEEE Trans. Multimedia}, vol. 27, pp. 1783--1796, 2025, doi: 10.1109/TMM.2024.3521798.

\bibitem{ref24}
Z. Chen, Z. He and Z. -M. Lu, "DEA-Net: Single Image Dehazing Based on Detail-Enhanced Convolution and Content-Guided Attention," in \textit{IEEE Trans. Image Process.}, vol. 33, pp. 1002--1015, 2024, doi: 10.1109/TIP.2024.3354108.

\bibitem{ref31}
Q. Huynh-Thu and M. Ghanbari, "Scope of validity of PSNR in image/video quality assessment," \textit{Electron. Lett.}, vol. 44, no. 13, pp. 800--801, 2008, doi: 10.1049/el:20080522.

\bibitem{ref25}
Z. Tu \textit{et al.}, "MAXIM: Multi-Axis MLP for Image Processing," in \textit{Proc. IEEE/CVF Conf. Comput. Vis. Pattern Recognit. (CVPR)}, New Orleans, LA, USA, 2022, pp. 5759--5770, doi: 10.1109/CVPR52688.2022.00568.

\bibitem{ref26}
S. Yamashita and M. Ikehara, "Image Deraining with Frequency-Enhanced State Space Model," in \textit{Proc. Asian Conf. Comput. Vis. (ACCV)}, vol. 15475, pp. 318--334, 2025, doi: 10.1007/978-981-96-0911-6\_19.


\bibitem{ref28}
L. Peng, A. Jiang, Q. Yi and M. Wang, "Cumulative Rain Density Sensing Network for Single Image Derain," in \textit{IEEE Signal Process. Lett.}, vol. 27, pp. 406--410, 2020, doi: 10.1109/LSP.2020.2974691.






\bibitem{ref33}
I. Loshchilov and F. Hutter, "SGDR: Stochastic gradient descent with warm restarts," arXiv:1608.03983, 2016.

\bibitem{ref35}
K. Jiang \textit{et al.}, "Multi-Scale Progressive Fusion Network for Single Image Deraining," in \textit{Proc. IEEE/CVF Conf. Comput. Vis. Pattern Recognit. (CVPR)}, Seattle, WA, USA, 2020, pp. 8343--8352, doi: 10.1109/CVPR42600.2020.00837.

\bibitem{ref36}
H. Wang, Q. Xie, Q. Zhao and D. Meng, "A Model-Driven Deep Neural Network for Single Image Rain Removal," in \textit{Proc. IEEE/CVF Conf. Comput. Vis. Pattern Recognit. (CVPR)}, Seattle, WA, USA, 2020, pp. 3100--3109, doi: 10.1109/CVPR42600.2020.00317.

\bibitem{ref37}
S. W. Zamir \textit{et al.}, "Multi-Stage Progressive Image Restoration," in \textit{Proc. IEEE/CVF Conf. Comput. Vis. Pattern Recognit. (CVPR)}, Nashville, TN, USA, 2021, pp. 14816--14826, doi: 10.1109/CVPR46437.2021.01458.

\bibitem{ref38}
X. Fu, Q. Qi, Z.-J. Zha, Y. Zhu, and X. Ding, “Rain Streak Removal via Dual Graph Convolutional Network”, in \textit{Proc. AAAI Conf. Artif. Intell. (AAAI)}, vol. 35, no. 2, pp. 1352--1360, May 2021.

\bibitem{ref39}
Q. Yi, J. Li, Q. Dai, F. Fang, G. Zhang and T. Zeng, "Structure-Preserving Deraining with Residue Channel Prior Guidance," in \textit{Proc. IEEE/CVF Int. Conf. Comput. Vis. (ICCV)}, Montreal, QC, Canada, 2021, pp. 4218--4227, doi: 10.1109/ICCV48922.2021.00420.

\bibitem{ref40}
J. Xiao, X. Fu, A. Liu, F. Wu and Z. -J. Zha, "Image De-Raining Transformer," in \textit{IEEE Trans. Pattern Anal. Mach. Intell.}, vol. 45, no. 11, pp. 12978--12995, Nov. 2023, doi: 10.1109/TPAMI.2022.3183612.

\bibitem{ref41}
H. Sun, J. Xu, J. Wang, Q. Qi, C. Ge and J. Liao, "DLI-Net: Dual Local Interaction Network for Fine-Grained Sketch-Based Image Retrieval," in \textit{IEEE Trans. Circuits Syst. Video Technol.}, vol. 32, no. 10, pp. 7177--7189, Oct. 2022, doi: 10.1109/TCSVT.2022.3171972.

\bibitem{ref42}
Z. Wang, X. Cun, J. Bao, W. Zhou, J. Liu and H. Li, "Uformer: A General U-Shaped Transformer for Image Restoration," in \textit{Proc. IEEE/CVF Conf. Comput. Vis. Pattern Recognit. (CVPR)}, New Orleans, LA, USA, 2022, pp. 17662--17672, doi: 10.1109/CVPR52688.2022.01716.



\end{thebibliography}

\end{document}